\documentclass{bmvc2k}

\usepackage{epsfig}
\usepackage{graphicx}
\usepackage{amsmath}
\usepackage{amssymb}

\title{Monocular Navigation in Large Scale Dynamic Environments}

\addauthor{Darius Burschka}{http://www6.in.tum.de/burschka/}{1}
\addinstitution{Department of Computer Science\\
Technische Universit\"at M\"unchen\\ Germany}

\runninghead{Darius Burschka}{Monocular Navigation in Dynamic Environments}

\begin{document}
\maketitle

\begin{abstract} We present a processing technique for a robust
reconstruction of motion properties for single points in large scale,
dynamic environments. We assume that the acquisition camera is moving
and that there are other independently moving agents in a large
environment, like road scenarios. The separation of direction and
magnitude of the reconstructed motion allows for robust reconstruction
of the dynamic state of the objects in situations, where conventional
binocular systems fail due to a small signal (disparity) from the images
due to a constant detection error, and where structure from motion
approaches fail due to unobserved motion of other agents between the
camera frames.

We present the mathematical framework and the sensitivity analysis for the resulting system.
\end{abstract}

\section{Motivation}
Depth information is essential for mobile systems to interact with the surrounding environment.  The reconstruction of static scenes is a well understood problem in {\em Computer Vision}, but most of the current applications need to cope with dynamic environments with multiple moving agents. The vision-based reconstruction systems can be categorized into binocular and monocular approaches that use the direct scene illumination or enhance the processing with active illumination patterns projected onto the scene.  Especially the later became very popular with the introduction of the PrimeSense sensor in the Microsoft Kinect camera.  The stereo approaches based on active illumination of the scene and binocular approaches suffer from the limitation in the achievable range. It depends on the brightness of the light source or the distance between the cameras that define the depth resolution of the system. Monocular approaches compensate this problem by providing a flexible distance between the acquired images used for 3D reconstruction. 
Depending on the velocity of the camera, the system can delay the acquisition of the second image based for example on the length of the sparse optical flow vectors from point correspondences between the images. 
In case of significant rotation during the acquisition, an additional compensation of the rotation in the optical flow may be necessary, because only the translational motion of the camera carries information about the depth. The problem with typical  monocular approaches is that they do not allow to reconstruct the scale for the reconstructed information and they fail to reconstruct the correct information in case of a motion in the scene between the two images used for reconstruction.  The motion estimation is usually done based on typical structure from motion approaches, like Essential or Homography matrix decompositions~\cite{Zisserman} or similar approaches. The achievable accuracy in case of static scenes is hereby strongly dependent on the distribution of the features in the images, which was derived in~\cite{elmarIros}.  They also require that the moving object provides at least 5~observed points to calculate the motion parameters~\cite{5points}.

While monocular approaches become very popular due to the compact system
dimensions and their scalabilty because of the flexible choice of the
baseline distance between the two camera images that are used for
reconstruction, they cannot cope with independent motion of structures
in the scene.  Some approaches treat these independently moving objects
as outliers, which can be analyzed later if they provide a sufficient
number of corresponding points on them or just discard them. The problem
is that since the 3D displacement of the structure is estimated here,
the rotational and translational component of the motion needs to be
reconstructed first. This information may be
very inaccurate, if the moving object covers only a very small part of
the image. The constant detection errors of the underlying points have a
large impact on the result, if the changes in the image are small.
Therefore, the estimation is reasonable for large objects in the images
and becomes useless for distant small objects (the 10 pixel problem in
pose detection of humans).

\vspace{-3ex} 
\subsection{Related Work}
There exist approaches to analyze the independent motion of clusters in the images. Approaches like the Generalized Principle Component Analysis (GPCA) can be used to find the independently moving clusters in images~\cite{Rene03}. There are approaches reconstructing the motion of moving objects in the world from multiple images between the images~\cite{Rene05,others}. In case of planar environments, the plane+parallax approach is applied to analyze the independent motion properties in the scene~\cite{ID11,ID12,ID28}.

Recently, multiple approaches have been published that combine Structure-from-motion and optical flow~\cite{ID33,ID29,ID31,ID20}. The current top method applied to the KITTI-2012 benchmark~\cite{ID33} calculates the fundamental matrix and computes the epipolar lines of the flow. This computation is limited to rigid scenes. A similar calculation based on fundamental matrix and regularization of the optical flow to align with the epipolar lines can be found in~\cite{ID31}. The independent motion in the scene is detected by reverting it to the optical flow of the entire scene.  Roussos~\cite{ID23} finds a solution for the depth and motion parameters for moving objects in the scene from batch processing on a sequence of about 30 frames.
There have been multiple approaches to motion segmentation of the scenes into regions corresponding to independently moving objects by exploiting 3D motion cues and epipolar motion~\cite{ID1,ID30,ID27}.

    This approach goes beyond  the problem of clustering of independent
    motion components. It provides a framework for  motion estimation in
    dynamic scenes using an extension of the Time-to-Collision Approach
    presented in~\cite{other}. It is interesting to see that under some
    restricted conditions, the system is able to reconstruct the depth
    relations entirely based on pixel information of single points in
    the images. The remaining paper is
    structured as follows. In Section 2, the new method of the depth
    calculation for point features in presented. In Section 3, the error
    propagation in the presented framework is presented We conclude with
    an evaluation of the achieved results.  

\section{Approach}

    Our proposed approach aims to estimate motion properties of single points in monocular image sequences. In case that both the camera and the object have independent motions in the environment, regular structure from motion approaches fail to reconstruct the motion and depth parameters correctly if only one point can be tracked on a distant object.

\begin{figure}[ht]
\begin{center}
\includegraphics[width=6.0cm]{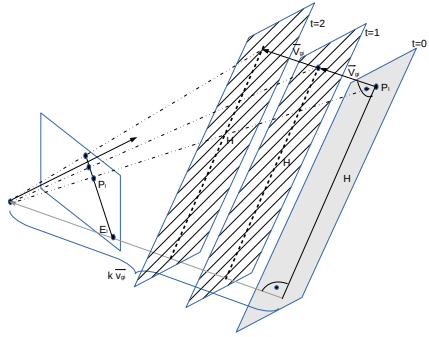}
\caption{\label{ttc:fig}A 3D point $\vec{P}_i$ moves in three frames
t={0,1,2} with a velocity vector~$\vec{v}_{gi}$. The
vector~$\vec{v}_{gi}$ together with the corresponding point~$\vec{P}_i$
defines a gray {\em collision plane} with~$\vec{v}_{gi}$ as its normal
vector.}
\end{center}
\end{figure}

It is obvious that relative motions of the camera~$\vec{v}_c$ and the motion of a point on an object~$\vec{v}_i$ in the environment are observed as an apparent combined motion vector~$\vec{v}_{gi}=\vec{v}_i-\vec{v}_c$.  The negative sign of~$\vec{v_c}$ is due to the apparent additional motion of the object due to the motion of the camera.  The sparse optical flow field reconstructed from point correspondences represents the projection of the resulting motion vector~$\vec{v}_{gi}$, which can be treated as if only the point~$p_i$ were moving in reference to a ``static'' camera.

Let us assume for now that a three-dimensional point~$\vec{P}_i$ moves with an arbitrary constant velocity vector~$\vec{v}_{gi}$ as depicted in Fig.~\ref{ttc:fig}. This motion results in a trace of projected points~$p_i$ in consecutive image frames for time steps t=\{0,1,2\}. The vector~$\vec{v}_{gi}$ normal to the plane containing~$P_i$ can be moved within this {\em collision plane} to align with the line going through the focal point of the camera.  The corresponding intersection point~$E_i$ of this line with the camera plane represents the projection of the observed point~$P_i$ at infinite distance from the camera~$-k\cdot\vec{v}_{gi}\rightarrow \infty$. Similar to the time-to-collision (TTC) approaches, the projected point~$p_i$ moves from the epipole~$E_i$ away along the line drawn in the camera image plane, while the 3D point~$\vec{P}_i$ moves closer to the camera in the scene.  

\begin{figure}[ht]
\begin{center}
\includegraphics[width=5cm]{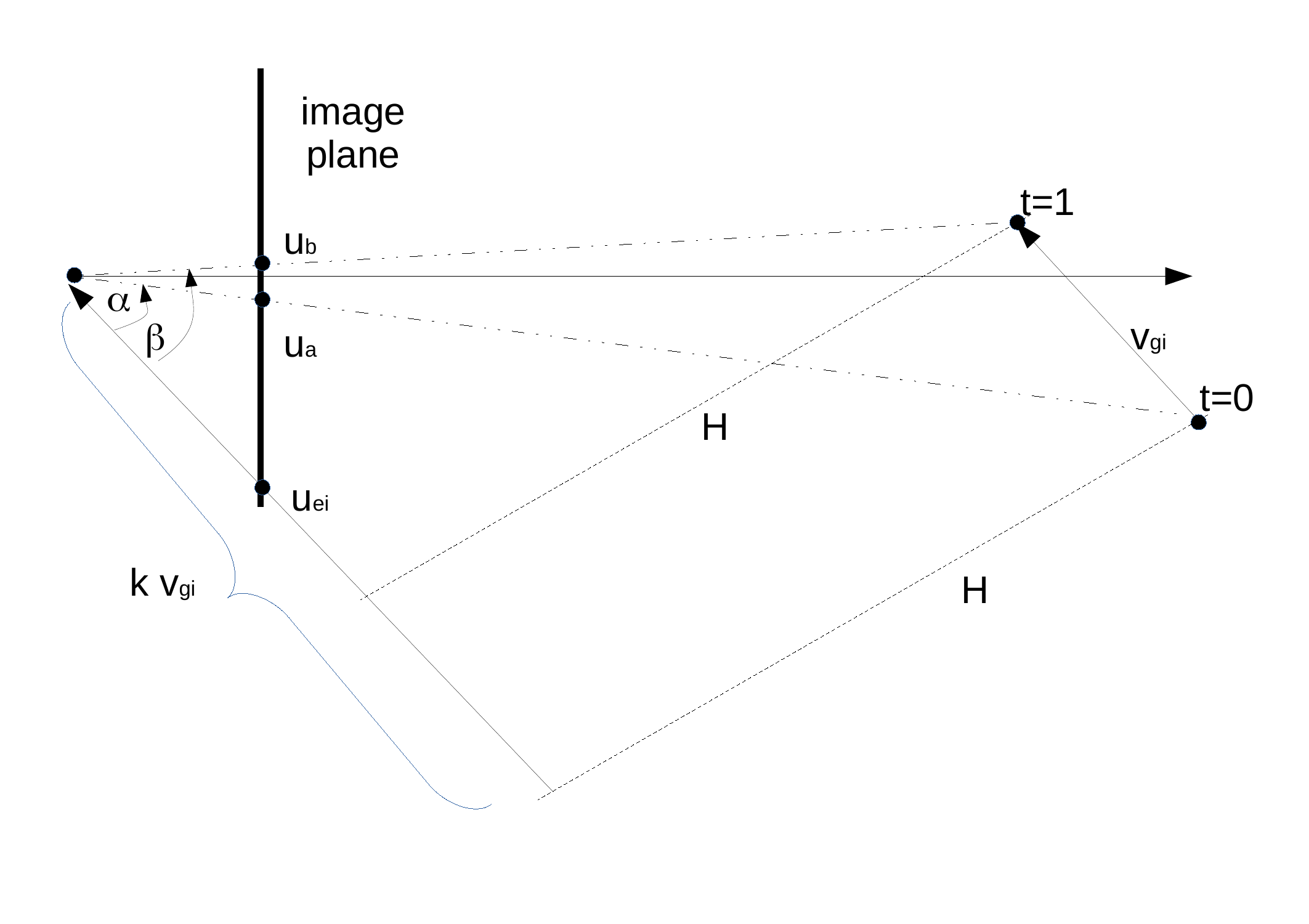}
\caption{\label{ttcplane:fig} Projection in the plane of the
set of projection triangles from Fig.~\ref{ttc:fig}.}
\end{center}
\end{figure}

For a safe navigation in a dynamic environment, it is often more important to understand the collision relations than the true metric distances to objects. So can a close object move in the same direction as the camera itself, while a distant object may be approaching the camera with a high speed. The measure for the relevance of a 3D point~$P_i$ can be replaced by estimating the time-to-collision (TTC) as the length~$k$ (Fig.~\ref{ttcplane:fig}), which tells the number of frames~k (k is a float number) until the {\em collision plane} containing the point~$P_i$ reaches the focal point of the camera. This value can be calculated directly from the image information without any additional extrinsic calibration data. We re-draw Fig.~\ref{ttc:fig} in a way that the set of projection triangles lies in the  image plane (Fig.~\ref{ttcplane:fig}).  A projection point~$p_i=(x_i,y_i,f)^T$ with f being the focal length in pixels can be converted into an angle relative to the optical axis (Fig.~\ref{ttcplane:fig}) first (\ref{angl:eq}):

\begin{eqnarray}
\label{angl:eq}
\gamma_i=\arctan\frac{u_i}{f},\quad
\alpha=\gamma_{u_a}-\gamma_{u_{e_i}}, \quad
\beta=\gamma_{u_b}-\gamma_{u_{e_i}}
\end{eqnarray}
The value k is reduced by 1 with each new frame.  Therefore, we can write the following equation (\ref{ttc:eq}):
\begin{eqnarray}
\nonumber
  \tan \alpha=\frac{H}{k\cdot||\vec{v}_{gi}||}, \quad 
  \tan \beta=\frac{H}{(k-1)\cdot||\vec{v}_{gi}||}, \quad
  \tan\alpha\cdot k\cdot||\vec{v}_{gi}||=\tan\beta\cdot
  (k-1)\cdot||\vec{v}_{gi}||\\[1.5ex]
  \Rightarrow k=\frac{\tan\beta}{\tan\beta-\tan\alpha}
  \label{ttc:eq}
\end{eqnarray}\vspace{-2ex}

\subsection{Simplified Motion Cases}

\begin{figure}[ht]
\begin{center}
\includegraphics[width=4cm]{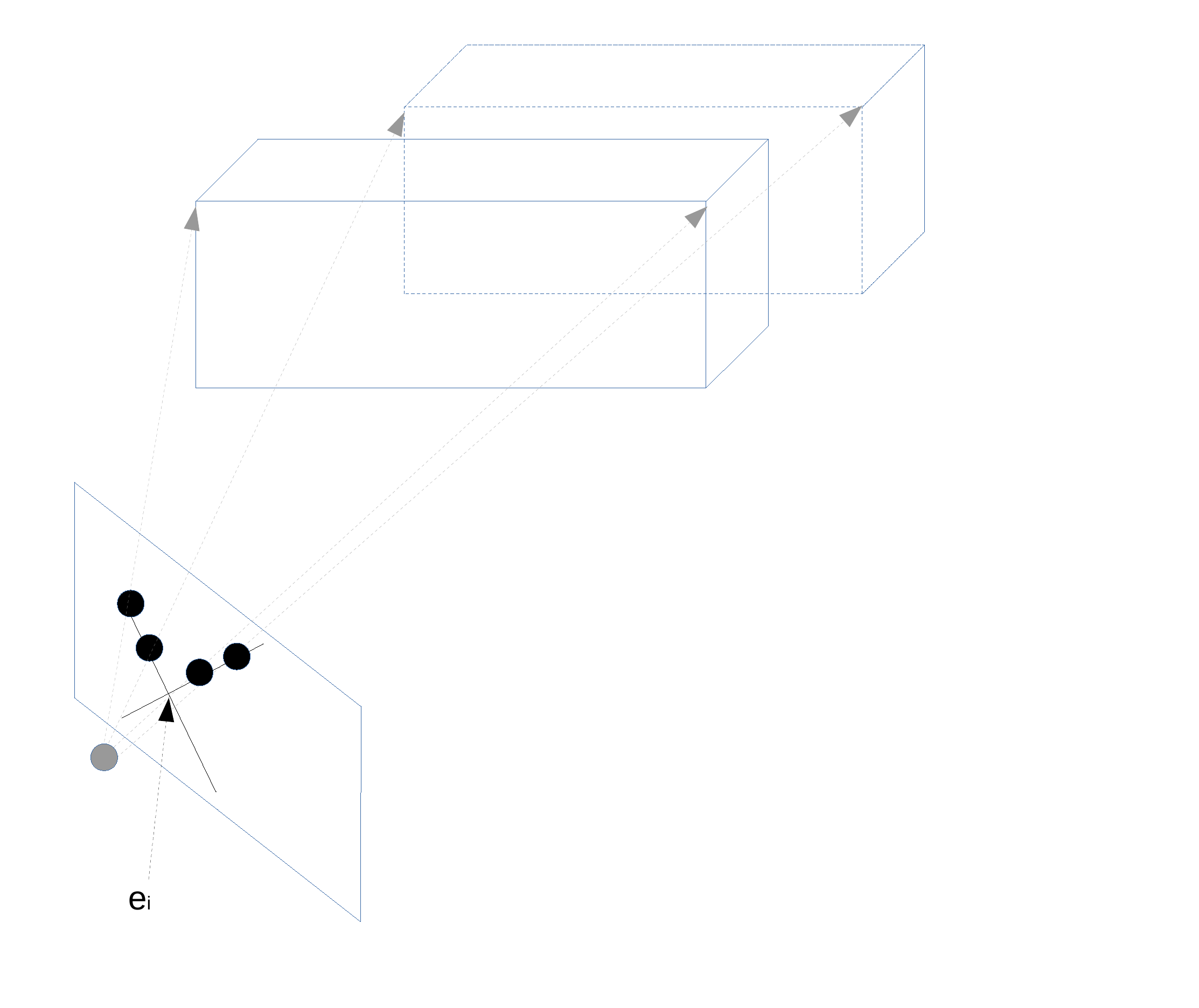}
\includegraphics[width=4cm]{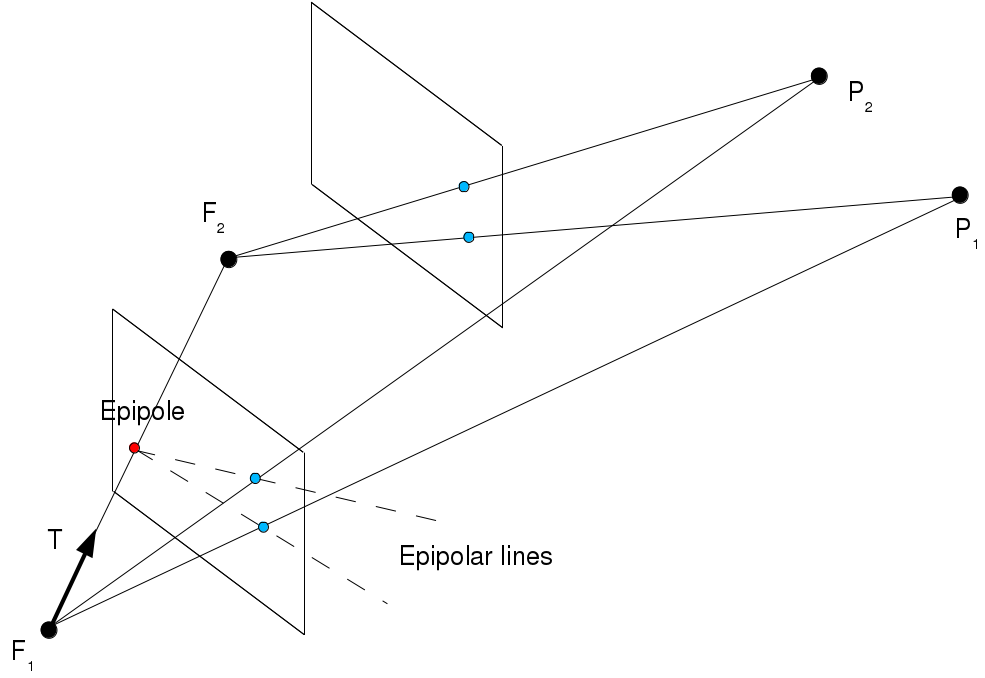}
\caption{\label{epiinter:fig} Epipole~$e_i$ can be found from the
intersection of the optical flow lines of at least two points on a rigid
object, iff there is no rotation involved.}
\end{center}
\end{figure}

\paragraph{Planar Motion} -
\label{sec:planar_motion}
The problem with equation (\ref{ttc:eq}) is the missing knowledge about the epipole position~$E_i$ for a given observed motion of a point~$p_i$. There are special cases for which the value~k can be estimated directly from the motion of a single point in two images. We can tell directly from Fig.~\ref{ttcplane:fig} that for all motions entirely in a horizontal plane, e.g. on the office floor or road, the epipole must lie on the horizon line in the image. This follows directly from the requirement that the motion vector~$v_{gi}$ needs to be moved in the plane until it goes through the focal point, which is a point on the horizon. Since there is no vertical component of the velocity for this case, the epipole needs to be in the horizontal plane of the focal point. If the observed point  has some height above the ground that moves it away from horizon line then the position of the epipole~$e_i$ can be found as the intersection of the optical flow line with the horizon line. The intersection point becomes increasingly more accurate, the further the observed point is from the horizontal plane, which means the further the evaluated 3D point is from the ground or the closer the camera is to the observed object.
The image position of~$e_i=(u_e,v_e)^T$ defines the direction of the motion vector to be along the vector $v_{gi}=(u_e,v_e,f)^T$ with f being the focal length in pixels according to the definition in Fig.~\ref{ttc:fig} and Fig.~\ref{epiinter:fig}. We can estimate the distance value k with equation~(\ref{ttc:eq}). As usual in structure from motion approaches, we cannot estimate the absolute motion value. The "depth" is composed from the  number of frames~k until the {\em collision plane} of the point~$P_i$ reaches the focal point of the camera and the corresponding orthogonal value H (Fig.~\ref{ttcplane:fig}) to (\ref{depth:eq}):
\begin{equation}
  H=k\cdot \tan\alpha, \quad
  \vec{v}_{gi}=\frac{\vec{e}_i}{||\vec{e}_i||},\quad
  \vec{v}_H=\frac{(\vec{v}_{gi}\times \vec{p}_i)\times\vec{v}_{gi}}
         {||(\vec{v}_{gi}\times \vec{p}_i)\times\vec{v}_{gi}||}\quad
\label{depth:eq}
  \Rightarrow\quad\quad\vec{P}'_i=k\cdot\vec{v}_{gi}+H\cdot\vec{v}_H
\end{equation}

The expression~$\vec{v}_H$ is a virtual shift within the {\em collision
plane} that allows to find the displacement factor~H that tells us, how
far the point is from the direct collision with the focal point.  All
distance parameters are expressed in {\em collision times} TTC.  It is
interesting to see that for an in-plane motion in a static environment,
the system can estimate the motion relative to any object from merely
pixel data of one point on the object.  In dynamic scenes, the notion of
depth is replaced by the notion of how soon a {\em collision plane}
including the point passes through the focal point of the camera. The
values~(k,H) describe when and in what distance from the focal point the
``collision'' will occur.  


\vspace{-3ex}
\paragraph{Direct Collision Candidate} -
Another special case exists, when the observed point is directly in the epipole~$e_i$ of the current motion. In this case, the distance H goes to zero (H=0). The point will always appear at the same position in the camera image. It becomes a static target. It is known in nautical applications as "constant bearing" often used to define collision candidates based on their apparent fixation at a specific angle (bearing) to the ship. The TTC expression for this point can only be calculated from the surrounding region around it.

\vspace{-3ex}
\paragraph{Multiple Points on a Translating Object}-
The epipole~$e_i$ of an object can be understood as the point that will represent the object at very large distances from the camera. While the object approaches the camera with a given velocity~$v_{gi}$, the points on the object expand from the epipole outwards. The direction of motion relative to the epipole can also be used to decide if the object approaches or escapes the camera. In the second case, the "collision" already happened. If the points of an object move away from the epipole then the collision is going to happen in~k or it happened already k-frames ago in other case. We can find the position of the epipole in that case using the fact that all points on a rigid structure share the same epipole. The line segments defined by at least two points on the object will intersect exactly in the epipole~$e_i$ (Fig.~\ref{epiinter:fig} left). Fig.~\ref{epiinter:fig} right shows that the direction of the relative motion between the camera and the object~$\vec{T}$ can be obtained directly from the position of the epipole. $(F_1,F_2)$ are the corresponding focal points for the two camera images of the sequence if we consider the relative motion to be performed by the camera.

For a set of corresponding points between two images with $\{\vec{p}_i\}$ being 2D image points in the first image and $\{\vec{p'}_i\}$ being corresponding image points in the second image, we can calculate the epipole~$e_x$ to~(\ref{epip:eq}).
\begin{equation}
 \label{epip:eq}
  \begin{matrix}
 \vec{t}_i=\vec{p'}_i-\vec{p}_i=(t_{ix}, t_{iy})^T,\quad
 \vec{n}_i=\frac{(-t_{iy},t_{ix})^T}{||(-t_{iy},t_{ix})^T||}, \quad
 \tilde{A}=\left(\begin{matrix}
               \vec{n}_1^T\\
	       \vec{n}_2^T\\
	       \dots\end{matrix}\right), \quad
	       \vec{b}=\left(\begin{matrix}\vec{p}_1^T\vec{n}_1\\ 
	              \vec{p}_2^T\vec{n}_2\\\dots
	       \end{matrix}\right),\\
	       \Rightarrow \quad\vec{e}_x=\tilde{A}^{-*}\cdot\vec{b}
\end{matrix}
\end{equation}
The optical flow vector~$\vec{t}_i$ is calculated from  point
correspondences. Switching and negating entries of the vector result in
a normal vector~$\vec{n}_i$ orthogonal to the optical flow line from the
current point correspondence. We stack all the possible matches from a
rigid object in a matrix~$\tilde{A}$. The pseudo-inverse calculation
of~$\tilde{A}^{-*}$ allows to calculate a least-square fit for the
epipole~$e_x$ to the current set of lines.\vspace{-2ex}

\subsection{Clustering of Multiple Independent Motion Groups} 
For
the case that we observe only translational motion of multiple objects including the camera, the motion clustering can be done based on the epipole estimation described in the previous paragraph. We use RANSAC to pick sets of point correspondences and calculate a hypothesis for the epipole using~(\ref{epip:eq}). Then we check all the normals to optical flow vectors~$\vec{n}_i$ in~(\ref{epip:eq}) for the distance of the corresponding line to the epipole using~(\ref{dist:eq}):
\begin{equation}
\label{dist:eq}
\begin{matrix}
\vec{l}_i=\vec{e}_x-\vec{p}_i, \quad
\Rightarrow \quad d_i=\left|\vec{n}_i^T\cdot\vec{l}_i\right| < \epsilon
\end{matrix}
\end{equation}
We calculate the projection of the connection vector~$\vec{l}_i$ between
the epipole~$\vec{e}_x$ of the cluster and a point on the flow
field~$\vec{p}_i$. The calculated distance~$d_i$ needs to be smaller
than an epsilon distance to allow for detection errors in the camera
image. We require also a similar TTC value for all the points of the
cluster. All vectors that have a distance to the line smaller than an
epsilon a grouped to one motion cluster and the remaining lines are ran
through the RANSAC process iteratively again. We require that a cluster
should have at least three line segments to be grouped together, because
any two non-parallel lines intersect somewhere.  Fig.~\ref{fig:rewe}
shows an example of clustering of objects based on their
epipoles.\vspace{-2ex}

\subsection{Finding Epipole for an Arbitrary Translational  Motion}

\begin{figure}[ht]
\begin{center}
\includegraphics[width=5.0cm]{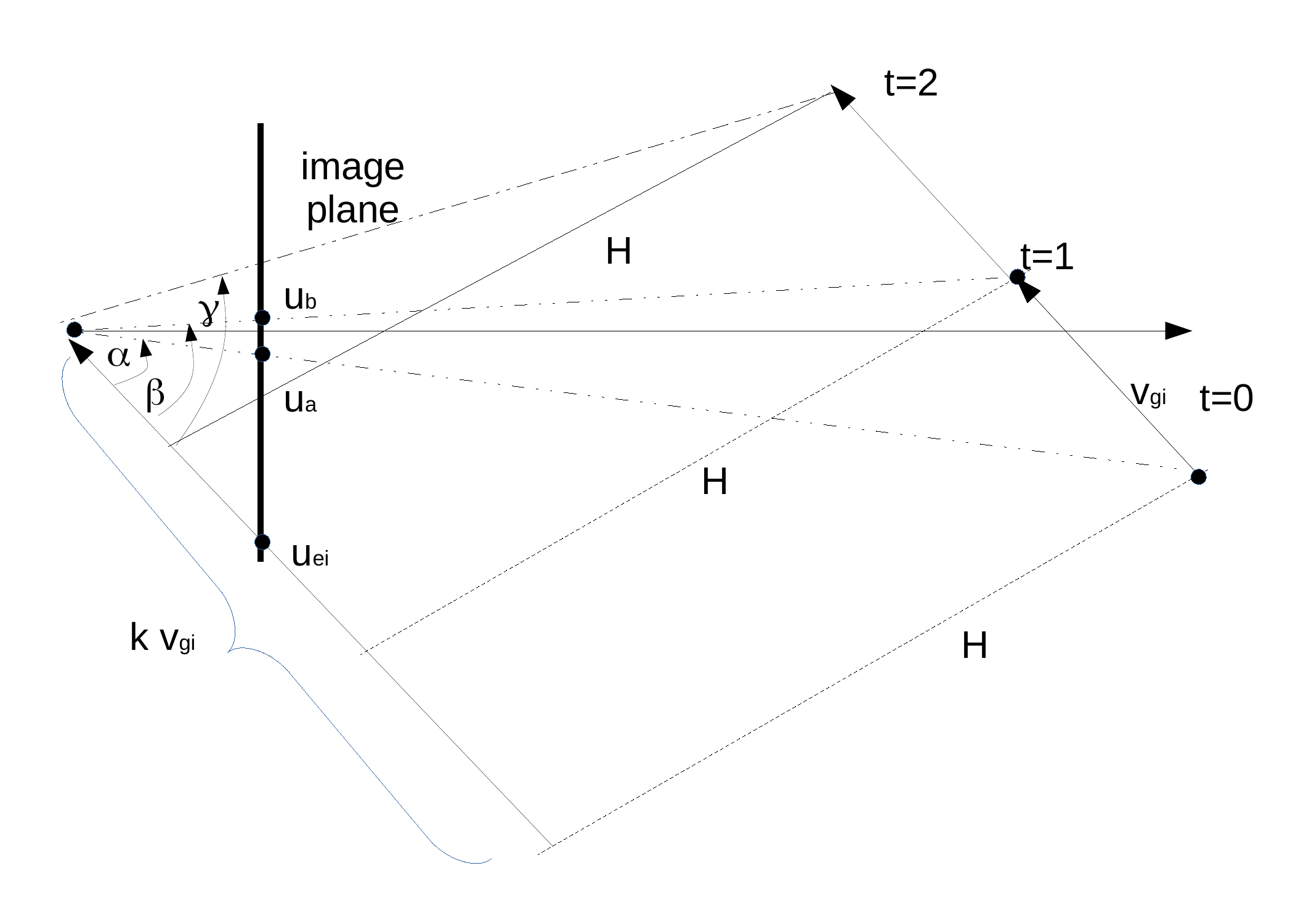}
\caption{\label{epiarb:fig} In a general motion case the epipole~$e_i$ needs
to be found from the position of a point tracked in three frames.}
\end{center}
\end{figure}

In case that the system observes a generic unconstrained motion in 3D space, the system needs to estimate the exact position of the epipole~$e_i$ along the optical flow line that is constructed from point positions in consecutive images. In a generic case, the epipole~$e_i$ in the image does not need to be on the horizon as it is the case for planar motion. In this case, an additional information from a third image is needed to estimate the position of the epipole from the change in the angular distances between the points (Fig.~\ref{epiarb:fig}). 

Our solution is to introduce an additional angular offset value~x, which is added to the horizontal case with the angles~$(\alpha,\beta)$ found from the intersection of with the horizontal line in the image as in the {\em Planar Motion Case} above. The true position is along the line away from the horizon shifted by an angle~x.  We can solve for this value by calculating the time to collision~k in~(\ref{ttc:eq}) for 2 consecutive frames of the sequence. The resulting difference between these values needs to be 1 since we reduced the number of frames by one. The solution for x is shown in~(\ref{xsolver:eq}):

\begin{equation}
\label{xsolver:eq}
\begin{matrix}
\frac{\tan(\beta+x)}{\tan(\beta+x) - \tan(\alpha+x)}-
\frac{\tan(\gamma+x)}{\tan(\gamma+x) -\tan(\beta+x))}\doteq1, \quad
\Rightarrow\quad
x=\arctan\frac{\tan\alpha\cdot\tan\beta - 2\tan\alpha\cdot\tan\gamma +
\tan\beta\cdot\tan\gamma}{\tan\alpha-2\tan\beta + \tan\gamma}
\end{matrix}
\end{equation}
We used the MATLAB symbolic solver to find this solution for x. We see
that all angular values~$\{\alpha,\beta,\gamma\}$ representing the
Planar Case have a constant correction value x that corrects for the
true position of the epipole relative to the horizon line. We define~$\{\alpha,\beta,\gamma\}$ in identical way as for the {\em Planar Motion Case} but need to add a correction value~x that needs to be estimated from the evolution of the angular properties over three frames. This has a disadvantage that we need to assume that the motion between the frames remains constant over the period of three frames.

\section{Experimental Results}

We implemented the framework on a Linux system using AKAZE features from OpenCV to estimate the sparse optical flow in the images. We can estimate the correspondences between the images with 12Hz, which defines the time-base for our TTC calculations. A comparison with an existing benchmark database was not possible here, since databases like KITTY do not provide any motion data for the moving objects in the scene, which is done by this framework.  An analysis of the sensitivity of the mathematical framework is provided here instead due to the lag of an appropriate ground-truth for comparisons.

\subsection{Accuracy of the Parameter Estimation compared to a
Bionocular Stereo Approach}

Most systems estimate motion parameters (magnitude,direction) from consecutive reconstructions of the depth for a point $P_i(t)=(X_i(t),Y_i(t),Z_i(t))^T$ on an object for time-stamps t={0,1}.  While this provides a valid solution for indoor applications in service robotics, we need to consider that the depth in binocular stereo is estimated from the horizontal projection change between the two cameras $d(t)=x_L(t)-x_R(t)$ to:
\begin{equation}
d(t)=\frac{B\cdot f}{Z_i(t)}
\end{equation}
with decreasing values of d(t) with increasing distance from the camera~$Z_i(t)$. Given a constant detection accuracy for the matched features, the error become very soon in the range of the expected disparity value~$d_i$ causing a large variation of the reconstructed point. The error gets propagated to the $(X_i(t),Y_i(t))$ values through the perspective projection equations.

For simplicity, let us compare the results with the planar case, where
epipole position in the presented framework can be calculated directly
from the intersection line of the flow segment with the horizon. The accuracy of direction depends on the accuracy of the intersection between the horizon and the line segment. The accuracy of the orientation of the segment increases with its length hence the detection accuracy can be assumed constant. We obtain following result for an object approaching with 50km/h with the motion direction of~$45^\circ$, with a baseline of the stereo system of 15cm, focal length=8mm, and detection error $\Delta p=0.2$pixels.

We can see that the proposed framework gives still useful results even at high distance Z from the camera, where the reconstruction error from binocular stereo renders the data useless.  
\begin{figure}[ht]
\begin{center}
\includegraphics[width=5.5cm]{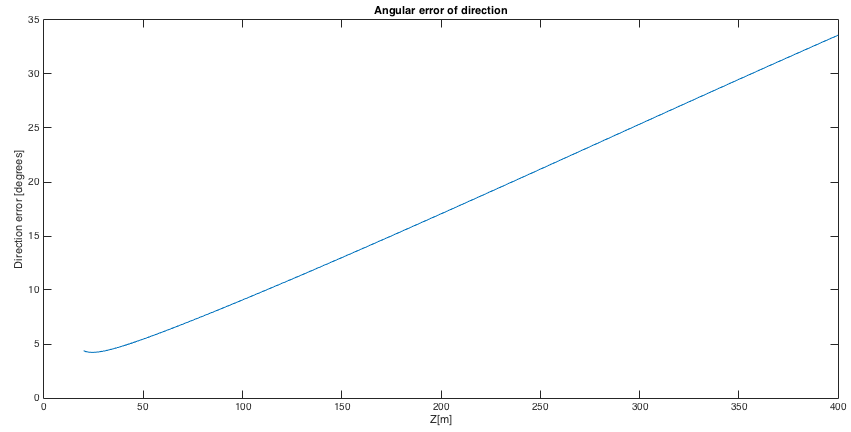}
\includegraphics[width=5.5cm]{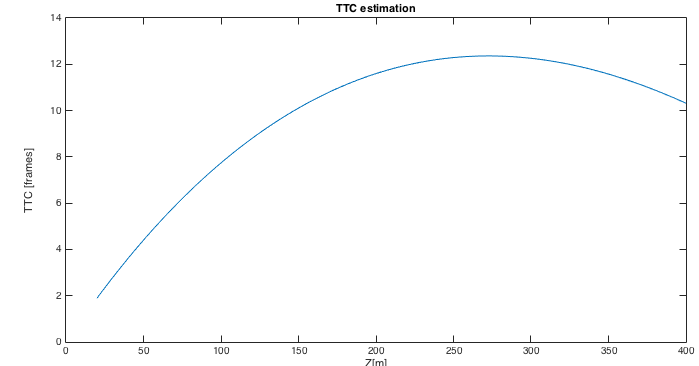}
\caption{\label{fig:motion_error}  Orientation error and TTC estimates
for the above example from the ``collision planes''  approach.}
\end{center}
\end{figure}

\vspace{-5ex}
\subsection{Planar Motion Examples}

\begin{figure}[ht]
\begin{center}
\includegraphics[width=5cm]{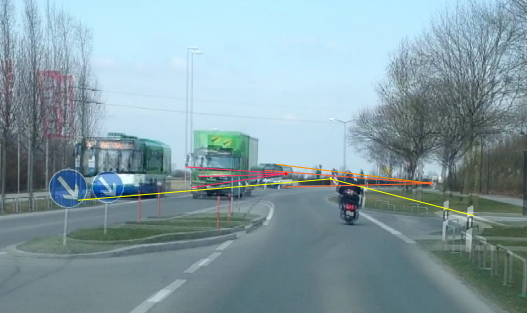}
\includegraphics[width=5cm]{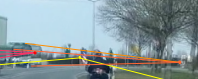}
\caption{\label{fig:rewe}Dynamic scene with 3 exemplary motion
components.
The yellow epipole shows the direction of the 
eigen-motion of the camera, the beige epipole is the direction
of motion for the bus in the circle, and the red epipole is the
direction of motion for the truck (other deleted to reduce confusion).}
\end{center}
\end{figure}

In case of a planar motion in the road scenario, we can drop the constraint on no rotation of the observed object that was required in Fig.~\ref{epiinter:fig}. For the planar motion in the examples of this subsection, all motion vectors have only horizontal (x,z)-components and, therefore, lie on the horizon line. This horizon line in the image is estimated in a calibration process, where a resting camera observes multiple linear motions of objects with different relative directions. In the calibration process, at least two points on the moving object are selected to find the epipole. The horizon can be found by connecting the estimated epipoles for different directions (Fig.~\ref{fig:rewe}). A possible error in the epipole position is corrected before applying the calculations for the TTC from Eq.~(\ref{ttc:eq}).

Fig.~\ref{fig:rewe} shows an example with multiple independent motion components in the image. For clarity, just the ego-motion of the camera (yellow), the motion direction of the truck (red) and the motion direction of the bus in the circle (beige) are shown with their epipoles in the image. A zoomed version shows the different motion directions represented as the direction to the epipole of the corresponding vectors. We took just a few vectors for each of them to avoid too much clutter in the image. We see, that the bus in the roundabout has a different moving direction than the truck avoiding the road divider and the biker in front of the car.

\subsection{Simulation Results}

\begin{figure}[ht]
\begin{center}
\includegraphics[width=4cm]{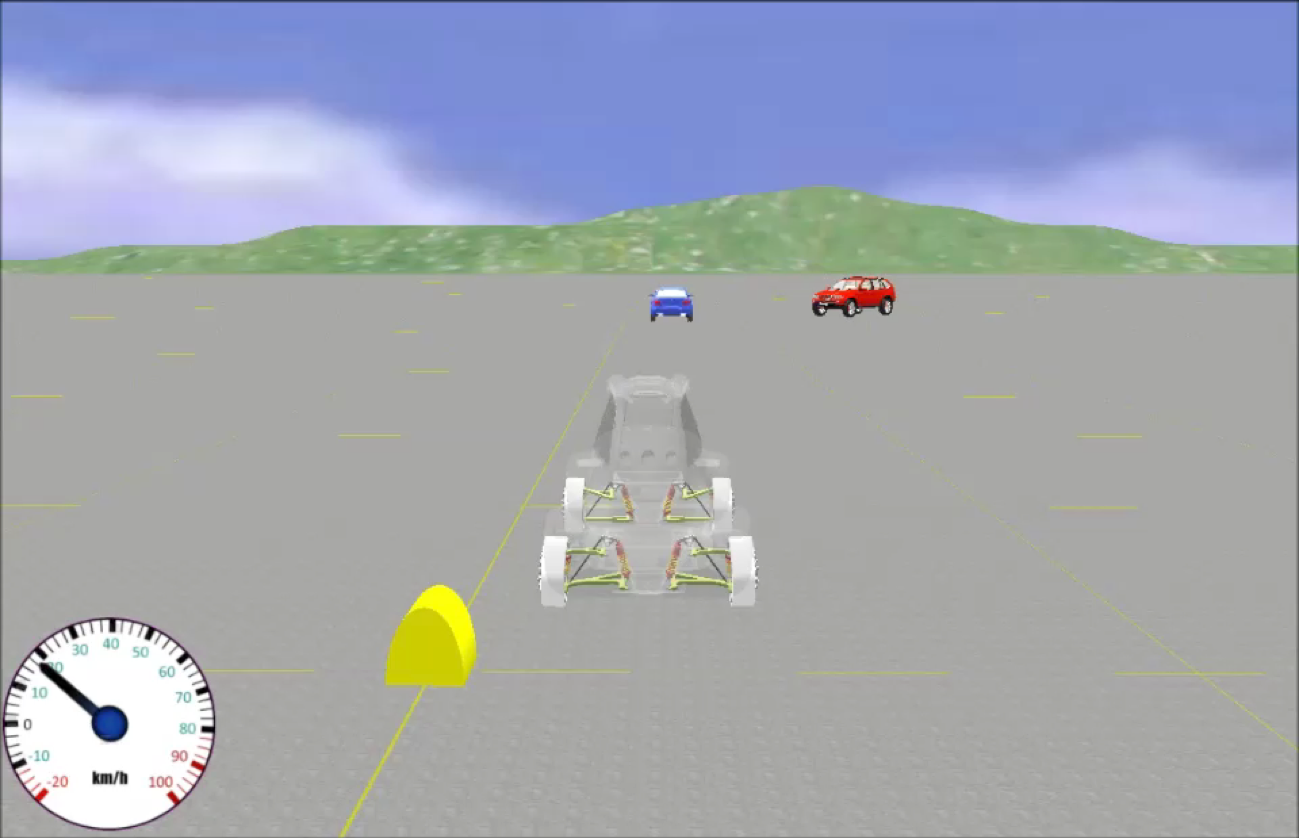}
\includegraphics[width=4cm]{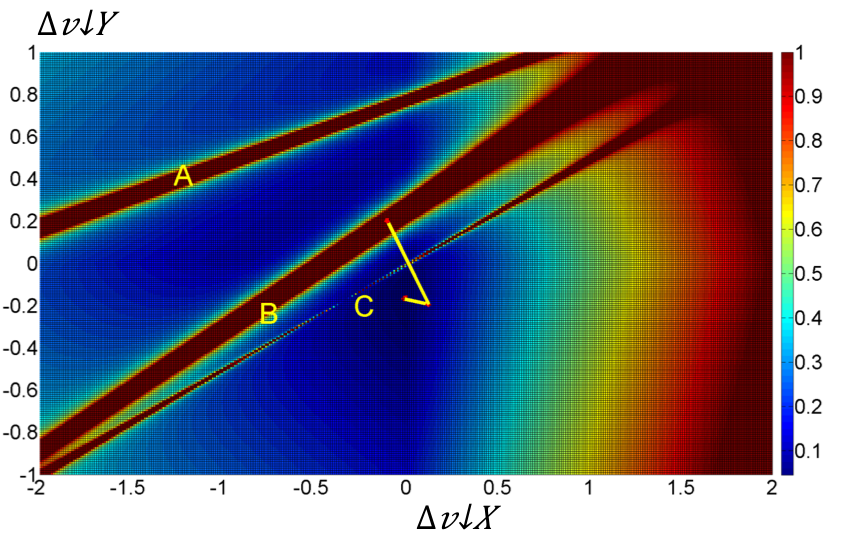}
\includegraphics[width=4cm]{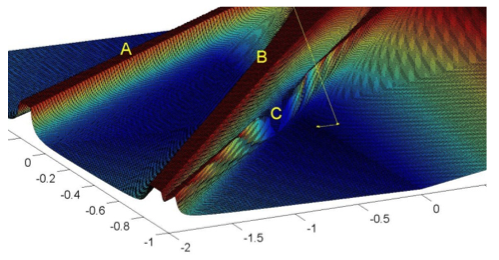}
\caption{\label{sim:fig} Collision map for a simulated scene on the top. The collision relations for different speed modifications of the own velocity in forwards and lateral directions are shown in the images below to be used for collision avoidance.}
\end{center}\end{figure}

While the presented approach is used on our car for low-level collision avoidance, the correctness of the approach was validated on simulated scenes, where collisions can easily be defined and validated, The system is able to calculate collision relations for the   dynamic scene with independently moving objects (Fig.~\ref{sim:fig}). The system can calculate the collision relations for different forwards and lateral velocity changes to the current vehicle speed (center of the planning map in Fig. \ref{sim:fig}bottom with $\Delta v=0$) and it allows to plan an evasion trajectory to avoid collisions.

\section{Conclusions and Future Work}

We presented a framework that allows to model the motion relations between moving objects in a dynamic environment. We used the collision properties of the planes including the observed point with the relative motion vector as a normal and their distance expressed as the time until this plane sweeps through the focal point of the camera to describe the motion properties of the scene. This formulation does not require any extrinsic calibration of the camera and allows a robust detection of collision candidates in front of the camera.  Since the direction of motion for the observed point is not calculated from position changes in the 3D~reconstructions of the scene in large distances from the camera, the resulting information is more reliable that the motion estimation from 3D readings. The modeling with the time-to-collision instead of metric distance as a parameter allows a better prioritization of collision candidates for the collision avoidance.  An object in 10m distance with the same motion direction as the camera does not create any danger to the camera while an object 75 meters away approaching the camera with high speed needs to be considered with a high priority. While a search for such object is difficult in Cartesian representations of the world, the distant  object will have a very small time-to-collision(TTC) while the TTC for the close object moving with the same speed and direction as the camera will approach infinity in our representation.  We use this framework in SenseAndAvoid applications for planes and as a safety system for collision avoidance in automotive applications. 

\small \bibliographystyle{egbib} \bibliography{lit}
\end{document}